\documentclass[9pt,twocolumn]{article}

\usepackage[
  letterpaper,
  top=57pt, bottom=73pt, left=54pt, right=54pt,
  columnsep=24pt,
  heightrounded
]{geometry}

\usepackage[utf8]{inputenc}
\usepackage[T1]{fontenc}
\usepackage{mathptmx}
\usepackage{microtype}

\usepackage{amsmath}
\usepackage{amssymb}
\usepackage{amsfonts}
\usepackage{graphicx}
\usepackage{booktabs}
\usepackage{caption}
\usepackage{adjustbox}
\usepackage{multicol}
\usepackage{multirow}
\usepackage{array}
\usepackage[dvipsnames,table]{xcolor}
\usepackage{placeins}
\usepackage{authblk}
\usepackage[colorlinks=true, linkcolor=Blue, citecolor=Blue, urlcolor=Blue]{hyperref}
\usepackage{cleveref}
\usepackage{cite}

\newcolumntype{C}{>{\huge}c}
\newcolumntype{L}{>{\huge}l}
\definecolor{mygray}{gray}{0.9}
\definecolor{lightblue}{RGB}{173,216,230}

\begin{document}

\title{\LARGE\bfseries HexMIL: Hierarchical Attention MIL for Ante-Hoc Explainable Detection of AI-Manipulated CT Volumes}

\author[1]{Orazio Pontorno}
\author[1]{Luca Guarnera}
\author[2]{Zahid Akhtar}
\author[1]{Sebastiano Battiato}

\affil[1]{Department of Mathematics and Computer Science, University of Catania, Catania, Italy}
\affil[2]{Department of Electrical and Computer Engineering, State University of New York Polytechnic Institute, Utica, NY, USA}

\date{}


\twocolumn[
  \begin{@twocolumnfalse}
    \maketitle
    \begin{abstract}
    The emergence of medical deepfakes, i.e., medical images manipulated by deep generative models, poses a significant threat to clinical workflows. However, existing detectors suffer from two critical limitations: poor generalization to unseen generative architectures for manipulation detection and lack of interpretability. In this context, we present \textit{HexMIL} (Hierarchical EXplainable Multiple Instance Learning), a mask-free medical deepfake detector that simultaneously addresses both limitations using only binary volume-level supervision. \textit{HexMIL} decomposes each CT volume into a two-level hierarchy of patches and slices, aggregated via independent Gated Attention modules whose weights are directly combined into a full-resolution 3D attention volume that localizes the manipulated sub-region without any pixel-level annotation. Unlike post-hoc methods such as Grad-CAM, \textit{HexMIL}'s attention weights constitute the exact forward computation driving the classification decision, providing ante-hoc and structurally faithful spatial attribution. We evaluate \textit{HexMIL} on M3DSynth and CT-GAN datasets under a rigorous cross-generator generalization protocol, training on a single generative architecture and testing on unseen ones. \textit{HexMIL} outperforms all baselines by $+9.1$ AUC and $+9.4$ F1 in out-of-domain classification, and achieves the best average IoU and Pointing Game score in localization. Project page: \href{https://opontorno.github.io/hexmil/}{opontorno.github.io/hexmil}.
    \end{abstract}
    \vspace{0.5em}
    \vspace{1em}
  \end{@twocolumnfalse}
]

\section{Introduction}
The integration of deep learning into medical imaging has led to significant advances in automated diagnosis and clinical decision support. However, these developments also introduce new security risks that remain largely unexplored. In particular, the emergence of medical deepfakes, namely medical images tampered by the use of deep generative models, affects the integrity of clinical workflows~\cite{li2025toward}. Recent advances in 3D generative models, such as Generative Adversarial Networks (GANs)~\cite{goodfellow2014generative} or Diffusion Models (DMs)~\cite{sohl2015deep, rombach2022ldm}, have allowed for the creation of volumetric scans with high visual fidelity, making such manipulations difficult to detect by both radiologists and standard diagnostic pipelines~\cite{dey2026advancements}. Moreover, real-world healthcare infrastructures typically use outdated protocols for managing data, such as Picture Archiving and Communication Systems (PACS)~\cite{solaiyappan2022machine}, which lack robust security mechanisms. In this context, an attacker could inject synthetic malignant lesions into healthy scans or remove existing pathologies, potentially leading to incorrect diagnoses and adverse patient outcomes~\cite{mirsky2019ct}. 
To contrast medical deepfakes, detector methodologies have been developed to identify the generative fingerprints left by local inpainting. Early approaches~\cite{solaiyappan2022machine} adapted Convolutional Neural Networks (CNNs) to classify 3D volumetric tampering, by detecting injections and removals of lung cancers generated by the pioneering CT-GAN framework~\cite{mirsky2019ct}. More recently, Zingarini~et~al~\cite{zingarini2024m3dsynth} explored the capability of recent object detectors to detect and localize tampered regions in AI-manipulated CT scans. However, since these technologies were not trained on medical data, they struggle in detecting and localizing manipulations. Other works focused on building robust  detectors using ad-hoc training strategies~\cite{dhanyalakshmi2026hybrid} or implementing more complex architectures~\cite{mahara2026forensic}. Moreover, unsupervised anomaly detection approaches have been proposed to detect medical deepfakes without using few labeled fake volumes. Back-in-Time Diffusion (BTD)~\cite{grabovski2025back} used the residual reconstruction error after a single reverse diffusion step, as metrics to localize synthetic content across CT and MRI scans. \\
Despite these advancements, current medical deepfake detectors still have trouble generalizing to unseen generative models, meaning they struggle to detect deepfakes generated by models that differ from those used to train the detectors~\cite{zingarini2024m3dsynth}.
Furthermore, all existing detectors are black boxes as they do not explain their decisions, resulting in a lack of transparency and interpretation. These characteristics are critical for the reliability of such technologies, especially when the results may affect a patient's treatment plan~\cite{pradepan2025comprehensive, kaleta2025jointdiffusion}. These issues highlight the urgent need for a medical deepfake detector that is robust to unseen generative architectures and natively interpretable without lacking accuracy in detecting AI-manipulated regions.\\
In this work, we propose \textit{HexMIL} (Hierarchical EXplainable Multiple Instance Learning), a mask-free framework that simultaneously addresses both limitations without requiring any pixel-level spatial supervision. Our method relies solely on binary labels (i.e., pristine or tampered), while segmentation masks are used exclusively for evaluation. First, \textit{HexMIL} hierarchically decomposes each CT volume into image patches extracted from axial slices. Grounded in Attention-Based Multiple Instance Learning~\cite{ilse2018attention}, \textit{HexMIL} leverages gated attention at patch and slice levels, named \textit{Patch Gated Attention} and \textit{Volume Gated Attention}, to improve detection performance and explainability. The two attention maps learned during training can be naturally combined into a full-resolution 3D attention volume that highlights the regions that contribute more to the model prediction, enabling localization of the manipulated sub-volume. Differently than methods like Grad-CAM~\cite{selvaraju2017grad} and Grad-CAM++~\cite{chattopadhay2018grad}, which rely on post-hoc gradient approximations, the attention weights directly contribute to the final prediction, resulting in more reliable explanation of the model decision.\\
We benchmark \textit{HexMIL} on two public datasets---M3DSynth~\cite{zingarini2024m3dsynth}, and CT-GAN~\cite{mirsky2019ct}. Specifically, we assess its capabilities in detection and localization performance within an out-of-domain context. While it is trained on tampered volumes generated by a single generative architecture, it is evaluated on volumes tampered by previously unseen architectures. Under this severe generalization setting, our method outperforms the state-of-the-art by a large margin.\\
\noindent The main contributions of this work are as follows:

\begin{itemize}

    \item \textbf{A unified framework for robust and interpretable 3D forensics.}
    \textit{HexMIL} is the first medical deepfake detector to jointly address out-of-domain generalization across unseen generative architectures, while being empowered with interpretrability.

    \item\textbf{Hierarchical mask-free training with intrinsic 3D spatial attribution.}
    \textit{HexMIL} does not require any pixel-level supervision, while still being able to detect and localize tampered regions.

    \item\textbf{Attention-based MIL for native 3D explainability without pixel-level supervision.}
    \textit{HexMIL} produces 3D attention volumes that highlight regions that contribute to the final prediction, providing inherent interpretability of the model decisions. To the best of our knowledge, this is the first application of attention-based multiple instance learning (MIL) for enhancing explainability in 3D medical deepfake detection.

    \item \textbf{State-of-the-art performance and cross-modality generalization.}
    \textit{HexMIL} achieves superior out-of-domain performance over all baselines, in both detection and localization. Furthermore, its intrinsic attention volumes outperform post-hoc methods such as Grad-CAM and Grad-CAM++ on both Pointing Game and IoU metrics, under identical supervision conditions.
\end{itemize}

The rest of the paper is structured as follows: \Cref{sec:related} presents a comprehensive review of the current state of the art in medical deepfake detection, alongside discussions on Multiple Instance Learning and Explainability in medical imaging. In \Cref{sec:method}, we provide an in-depth overview of \textit{HexMIL}'s architecture and training process. \Cref{sec:experiments} details the experimental results obtained in both out-of-domain classification and localization settings. Finally, \Cref{sec:conclusion} discusses the conclusions and limitations of the method.

\section{Related Work}
\label{sec:related}

\noindent\textbf{AIGC Image detection.}
The expansion of generative models has driven a rich body of forensic research on AI-generated content (AIGC). Early work concentrated on facial manipulation, exploiting low-level artifacts such as blending boundaries~\cite{rossler2019faceforensics++}, physiological inconsistencies~\cite{matern2019exploiting}, or classifier-based approaches trained on compression residuals~\cite{afchar2018mesonet}. As GAN synthesis matured, attention shifted to spectral signatures: CNNs trained for image generation leave characteristic high-frequency fingerprints that generalize across architectures~\cite{wang2020cnn, frank2020leveraging}. The challenge of detecting unseen generators then motivated universal, artifact-agnostic detectors~\cite{tan2024rethinking, yang2025d, pontorno2025deepfeaturex} and, more recently, methods tailored to the qualitatively distinct fingerprint of diffusion models, whose spectral profile differs markedly from that of GAN outputs~\cite{corvi2023detection, pontorno2024exploitation}.\\
\noindent\textbf{Medical image forensics.}
Mirsky~et~al.~\cite{mirsky2019ct} showed that a conditional GAN can inject or remove lung nodules from CT scans with quality sufficient to fool both radiologists and commercial CAD systems, establishing malicious CT tampering as a concrete clinical risk, giving rise to the phenomenon of Medical Deepfakes. On the detection side, Solaiyappan and Wen~\cite{solaiyappan2022machine} benchmarked standard 3D CNNs as binary classifiers, while MedNet~\cite{albahli2024mednet} added spatial-channel attention for improved sensitivity. MVSS-Net~\cite{chen2021image} pushed further toward spatial localization via multi-view, multi-scale supervision, though it requires dense pixel-level annotations that are prohibitively expensive for 3D volumes. ManTraNet~\cite{wu2019mantra} approaches the problem differently, using an LSTM-based anomaly scoring module pre-trained on a large catalogue of manipulation types; TruFor~\cite{guillaro2023trufor} extends this direction with a Transformer backbone that jointly exploits RGB appearance and a learned noise-sensitive fingerprint. Both achieve competitive results on 2D forensics benchmarks, yet process each CT slice independently and discard inter-slice volumetric context. More recently, Back-in-Time Diffusion~\cite{grabovski2025back} proposed an unsupervised alternative based on reverse-diffusion residuals; the approach works well for injection attacks but struggles with removals, where the inpainted region is statistically close to healthy tissue. The M3DSynth benchmark~\cite{zingarini2024m3dsynth} now provides a controlled test bed spanning three generative families (pix2pix, CycleGAN, Diffusion Model) for both attack types, exposing the brittleness of existing detectors when evaluated outside their training distribution. Our work departs from this supervised-unsupervised dichotomy: we rely only on volume-level binary labels, yet recover spatial localization as a by-product of the classification mechanism itself.\\
\noindent\textbf{Multiple Instance Learning.}
Multiple Instance Learning (MIL), introduced by Dietterich~et~al.~\cite{dietterich1997mil}, frames classification as a bag-level task over unordered instance sets. Deep MIL was transformed by Ilse~et~al.~\cite{ilse2018attention}, who replaced fixed pooling with a learnable gated attention mechanism that weights each instance by its estimated relevance, a formulation we adopt. In computational pathology, where gigapixel whole-slide images share the ``many patches, one label'' structure, attention MIL has become dominant: CLAM~\cite{lu2021data} combined instance-level clustering with attention for data-efficient classification; DSMIL~\cite{li2021dual} introduced dual-stream contrastive pretraining; TransMIL~\cite{shao2021transmil} injected Transformer-based correlation among instances; and DTFD-MIL~\cite{zhang2022dtfd} proposed double-tier feature distillation for pseudo-bag augmentation. However, all existing MIL architectures target 2D histology, processing a single tissue section as one bag. \textit{HexMIL} extends the paradigm to 3D CT volumes by introducing a second, independent aggregation level  (patch~$\to$~slice~$\to$~volume), with a two-stage training protocol that decouples low-level forensic feature learning from inter-slice evidence aggregation.\\
\noindent\textbf{Explainability in medical imaging.}
Post-hoc attribution methods remain the dominant approach to model interpretation in medical imaging. CAM~\cite{zhou2016learning} and its gradient-based extensions Grad-CAM~\cite{selvaraju2017grad} and Grad-CAM++~\cite{chattopadhay2018grad} generate saliency maps by backpropagating class-specific gradients through a frozen classifier. While widely adopted, these maps are approximations:\\ Adebayo~et~al.~\cite{adebayo2018sanity} showed that several saliency methods produce visually plausible but semantically meaningless maps that survive model-weight randomization, revealing a fundamental faithfulness gap. In contrast, the attention weights in \textit{HexMIL} are the exact aggregation function that produces the classification score --- they are not retrofitted onto a frozen model but are intrinsic to inference. Removing or randomizing them directly collapses the model to uniform averaging and measurably degrades classification, providing an intrinsic attribution mechanism that is structurally more aligned with the classification decision than any post-hoc approximation, since it constitutes the exact forward computation rather than a differentiable surrogate.

\begin{figure*}
    \centering
    \includegraphics[width=\linewidth]{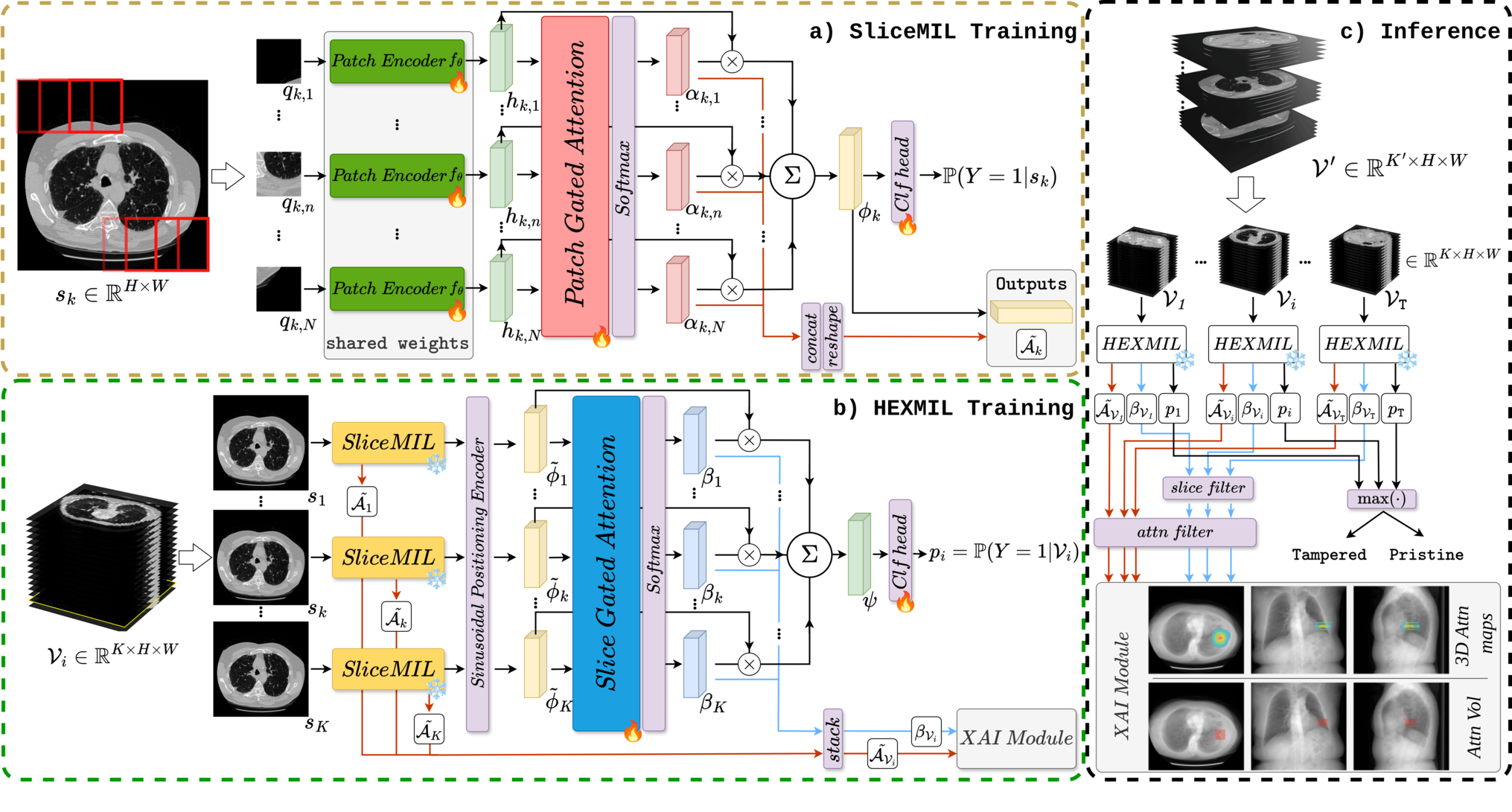}
    \vspace{-.6cm}
    \caption{
        \textit{HexMIL} graphical overview. 
        \textbf{(a) SliceMIL training:} We train SliceMIL encoder module to distinguish between pristine and tampered slices based on their features (\Cref{sec:slice}).
        \textbf{(b) HexMIL Training:} We train HexMIL model to identify and classify the presence of AI-manipulation within volume (\Cref{sec:volume}).
        \textbf{(c) Inference:} A generic K'-slices volume is divided into K-slices sub-volumes during testing, and HexMIL processes each one. The highest of them determines the final score (\Cref{sec:xai}).
        \includegraphics[height=10pt]{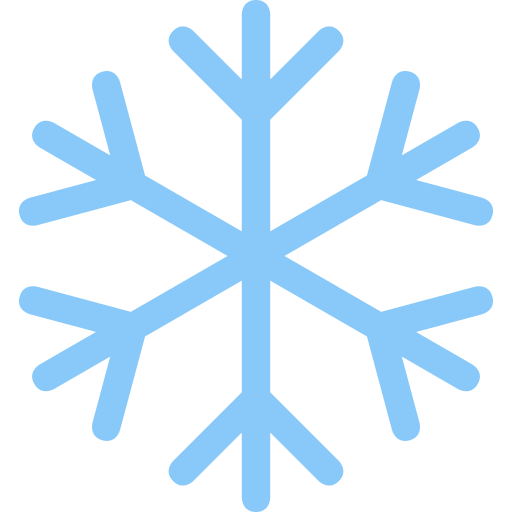} Module with frozen parameters. \includegraphics[height=10pt]{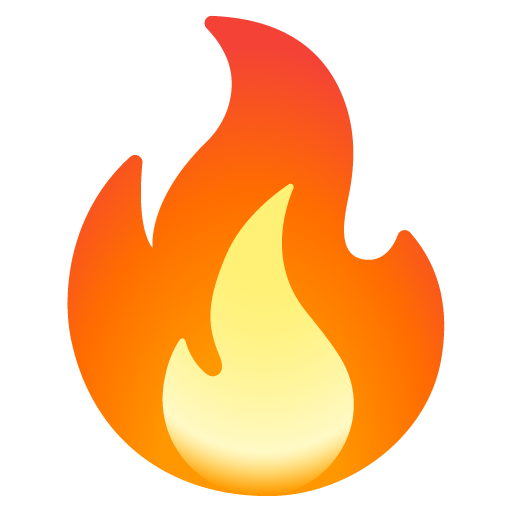} Module with trainable parameters.}
    \label{fig:pipeline}
\end{figure*}

\section{Proposed Method}
\label{sec:method}
The workflow of \textit{HexMIL} is depicted in \Cref{fig:pipeline}. The framework is trained in two sequential stages. In the first stage (\Cref{fig:pipeline}a), we train a 2D slice encoder, named SliceMIL, for a binary classification task. The aim is to produce a discriminative representation of input slices using a gated attention pooling mechanism~\cite{ilse2018attention}, which also provides an attention map focusing on the tampered region.
Once SliceMIL is trained, it is frozen and used as the slice encoder. In the second stage (\Cref{fig:pipeline}b), we train the entire \textit{HexMIL} architecture: an input volume of $K$ slices is decomposed into individual slices, each of which is processed by the SliceMIL encoder. The resulting representations are combined using a novel independent gated attention pooling mechanism, which generates a 1D attention vector highlighting the most likely tampered slices. This resulting volume-level feature vector is then processed by a classification head designed to determine whether the volume contains a tampered sub-volume.
\Cref{fig:pipeline}c illustrates the inference process. Since the model is trained on volumes of length $K$, and in real-world scenarios a CT scan may consist of a larger number $K'$ of slices, \textit{HexMIL} divides the entire volume into $T$ sub-volumes of length $K$ and assigns a probability to each sub-volume. The final probability assigned to the input volume is given by the maximum of these probabilities. Finally, the two attention maps are combined, yielding a 3D attention volume.

In the sections below, we provide a detailed problem formulation and a description of our method.

\subsection{Problem formulation}
\label{sec:probl}

We approach CT manipulation detection as a hierarchical binary classification problem. The task can be formalized as follows: let $\mathcal{W} \in \mathbb{R}^{K\times H\times W}$ be a CT scan volume space and consider a dataset $\mathcal{D}^V=\mathcal{P}^V\cup \mathcal{T}^V$, composed of pristine volumes $\mathcal{P}^V \subset \mathcal{W}$, and tampered volumes $\mathcal{T}^V \subset \mathcal{W}$. The tampered scans are generated by a set of $Q$ generators $\mathcal{G} = \{\mathcal{G}_i\}^{Q}_{i=1}$, for each $i$, $\mathcal{T}^V_i = \{t^i_1, \ldots, t^i_{m_i}\}$ denotes the $m_i$ volumes generated by $\mathcal{G}_i$, hence $\mathcal{T}^V = \bigcup_{i=1}^Q\mathcal{T}^V_i$. Given an input volume $\mathcal{V} \in \mathcal{W}$, the objective is to predict the binary label $\hat{y} \in \{0,1\}$ ($0$: pristine, $1$: tampered).

We partition the $Q$ generators into two disjoint sets for training and testing. The training set $\mathcal{D}^V_{train} = \mathcal{P}^V_{train} \cup (\bigcup_{i=1}^{\omega} \mathcal{T}^V_i)$ contains pristine volumes $\mathcal{P}^V_{train}$ and tampered volumes from $\omega < Q$ seen generators, while the test set $\mathcal{D}^V_{test} = \mathcal{P}^V_{test} \cup (\bigcup_{i=\omega+1}^{Q} \mathcal{T}^V_i)$ contains pristine volumes $\mathcal{P}^V_{test}$ and tampered volumes from the remaining $Q - \omega$ unseen generators, with $\mathcal{P}^V_{train} \cup \mathcal{P}^V_{test} = \mathcal{P}^V$ and $\mathcal{P}^V_{train} \cap \mathcal{P}^V_{test} = \emptyset$.

Finally, from $\mathcal{D}^V_{train}$ we derive a slice-level dataset $\mathcal{D}^{S}_{train} = \mathcal{P}^S_{train} \cup \mathcal{T}^S_{train}$, where pristine slices $\mathcal{P}^S_{train}$ are sampled randomly from pristine volumes, and tampered slices $\mathcal{T}^S_{train}$ are those intersecting the manipulated sub-volume, identified via the nodule axial coordinate available as standard dataset metadata. An analogous procedure yields $\mathcal{D}^{S}_{test}$.

\subsection{SliceMIL: Slice-Level Patch Aggregation}
\label{sec:slice}

In this stage, we explain the training process of the slice encoder, SliceMIL. Let $s_k \in \mathcal{D}^S_{train}$ be a training slice. This is decomposed into $N$ patches of size $P$ via a regular sliding window: $s_k = \{q_{k,1}, \ldots, q_{k,N}\}$, with $q_{k,n} \in \mathbb{R}^{P \times P}$, $n = 1, \ldots, N$. \\
\noindent\textbf{Patch encoding.}
Each patch $q_{k,n}$ is independently mapped to a $D$-dimensional feature vector by a shared CNN encoder $f_\theta$:
\begin{equation}
    h_{k,n} = f_\theta(q_{k,n}) \;\in\; \mathbb{R}^{D},
    \quad n = 1, \ldots, N,\;\; k = 1, \ldots, K,
    \label{eq:patch_enc}
\end{equation}
where $D$ is the feature dimension; $h_{k,n}$ is a compact representation of the local texture and structural patterns within each patch.\\
\noindent\textbf{Patch Gated Attention pooling.}
To aggregate the $N$ patch features of slice $s_k$ into a single slice representation $\phi_k$, we adopt the Gated Attention mechanism of Ilse~\cite{ilse2018attention}: 
\begin{equation}
    \alpha_{k,n} = \frac{
        \exp\!\Bigl(\mathbf{w}^{\!\top}
            \bigl(\tanh(\mathbf{V} h_{k,n})
            \odot\, \sigma(\mathbf{U} h_{k,n})\bigr)\Bigr)
    }{
        \displaystyle\sum_{j=1}^{N}
        \exp\!\Bigl(\mathbf{w}^{\!\top}
            \bigl(\tanh(\mathbf{V} h_{k,j})
            \odot\, \sigma(\mathbf{U} h_{k,j})\bigr)\Bigr)
    },
    \label{eq:alpha}
\end{equation}
where $\mathbf{V}, \mathbf{U} \in \mathbb{R}^{L \times D}$ are learnable projection matrices and $\mathbf{w} \in \mathbb{R}^{L}$ is a learnable weight vector, with $L = \dim(\alpha_{k,n})$. The $\tanh(\cdot)$ branch extracts signed feature activations, while the sigmoid gate $\sigma(\cdot) \in (0,1)$ selectively suppresses non-discriminative patches regardless of the sign of their activations~\cite{ilse2018attention}. The weights $\alpha_{k,n}$ sum to 1 over $n$ and are directly interpretable as the relative forensic relevance of each patch within the slice. The aggregated slice representation is:
\begin{equation}
    \phi_k = \sum_{n=1}^{N} \alpha_{k,n}\, h_{k,n} \;\in\; \mathbb{R}^{D}.
    \label{eq:slice_rep}
\end{equation}
A two-layer fully-connected head $g_\eta \colon \mathbb{R}^{D} \to \mathbb{R}$ produces a binary logit $\hat{\ell}_k = g_\eta(\phi_k)$ for slice $s_k$.\\
\noindent\textbf{Training.}
Parameters $\theta, \mathbf{V}, \mathbf{U}, \mathbf{w}, \eta$ are optimized jointly. No spatial supervision is applied within the slice: the gated attention mechanism must discover which patches are forensically relevant autonomously. The training objective is binary cross-entropy (BCE):
\begin{equation}
    \mathcal{L}_{\mathrm{slc}} = -\frac{1}{|\mathcal{D}^S_{train}|}
    \sum_{(s_k, y_k) \in \mathcal{D}^S_{train}}
    \Bigl[
        y_k \log \sigma\!\bigl(\hat{\ell}_k\bigr)
        + (1{-}y_k)\log\!\bigl(1 - \sigma\!\bigl(\hat{\ell}_k\bigr)\bigr)
    \Bigr],
    \label{eq:loss_cls}
\end{equation}

where $\hat{\ell}_k = g_\eta(\phi_k)$.

\subsection{HexMIL: Volume-Level Slice Aggregation}
\label{sec:volume}

After Stage 1 converges, the patch encoder $f_\theta$ and the entire SliceMIL sub-network $\{f_\theta,\mathbf{V},\mathbf{U},\mathbf{w},g_\eta\}$ are frozen and integrated into the \textit{HexMIL} model.\\
Let $\mathcal{V} = \{s_1, s_2, \ldots, s_K\} \in \mathcal{D}^V_{train}$ denote a CT volume consisting of $K$ axial slices. For each slice $s_k$ inside the volume we extract its representation $\phi_k = \textit{SliceMIL}(s_k)$. Obtaining the set of all $K$ slice representations $\{\phi_1, \ldots, \phi_K\}$.\\
\noindent\textbf{Sinusoidal positional encoding.} 
The slice encoder \textit{SliceMIL} processes each slice independently, meaning that the representations $\phi_k$ do not convey information about the specific location within the volume from which each slice was taken.\\
To maintain this spatial ordering, we incorporate a fixed sinusoidal  positional encoding (PE)~\cite{vaswani2017attention} into each slice representation prior to aggregating them at the volume level. The position-aware slice representation is:

\begin{equation}
    \tilde{\phi}_k = \phi_k + \mathrm{PE}(z_k) \;\in\; \mathbb{R}^{D}. \label{eq:pe_add}
\end{equation}

\noindent\textbf{Slice Gated Attention pooling.}
The position-aware representations $\tilde{\phi}_k$ are aggregated by a second independent Gated Attention module:
\begin{equation}
    \beta_k = \frac{
        \exp\!\Bigl(\mathbf{w}'^{\!\top}
            \bigl(\tanh(\mathbf{V}' \tilde{\phi}_k)
            \odot\, \sigma(\mathbf{U}' \tilde{\phi}_k)\bigr)\Bigr)
    }{
        \displaystyle\sum_{j=1}^{K}
        \exp\!\Bigl(\mathbf{w}'^{\!\top}
            \bigl(\tanh(\mathbf{V}' \tilde{\phi}_j)
            \odot\, \sigma(\mathbf{U}' \tilde{\phi}_j)\bigr)\Bigr)
    },
    \label{eq:beta}
\end{equation}
where $\mathbf{V}', \mathbf{U}' \in \mathbb{R}^{L' \times D}$, $L'=\dim(\beta_k)$, and $\mathbf{w}' \in \mathbb{R}^{L'}$ are independent from the slice-level parameters in Eq.~\eqref{eq:alpha}.
The volume representation and its classification logit are:
\begin{equation}
    \psi = \sum_{k=1}^{K} \beta_k\, \tilde{\phi}_k \;\in\; \mathbb{R}^{D},
    \qquad
    \hat{y} = \mathbf{1}\!\left[\sigma\!\bigl(h_\xi(\psi)\bigr) \geq 0.5\right],
    \label{eq:vol_rep}
\end{equation}
where $h_\xi \colon \mathbb{R}^{D} \to \mathbb{R}$ is a fully-connected classification head. The scalar $\beta_k$ quantifies how much each slice contributes to the volume-level decision; it forms the inter-slice component of the 3D attention heatmap (Section~\ref{sec:xai}).\\
\noindent\textbf{Training.}
With the slice-level parameters $\theta, \mathbf{V}, \mathbf{U}, \mathbf{w}, \eta$ frozen, only the volume-level parameters $\{\mathbf{V}', \mathbf{U}', \mathbf{w}', h_\xi\}$ are optimized:
\begin{equation}
\small
    \mathcal{L}_V = -\frac{1}{|\mathcal{D}^V_{train}|}
    \sum_{(\mathcal{V},\, y) \in \mathcal{D}^V_{train}}
    \Bigl[
        y \log \sigma\!\bigl(h_\xi(\psi)\bigr)
        + (1{-}y)\log\!\bigl(1 - \sigma\!\bigl(h_\xi(\psi)\bigr)\bigr)
    \Bigr].
    \label{eq:loss_vol}
\end{equation}
Freezing SliceMIL decouples patch-level representation learning from inter-slice evidence aggregation, preventing catastrophic forgetting of low-level forensic features while allowing $\beta_k$ to specialize on detecting which axial position carries the strongest manipulation signature. Neither $\mathcal{L}_{\mathrm{slc}}$ nor $\mathcal{L}_V$ employs any spatial supervision; the 3D attention structure that delineates the manipulated region emerges entirely as a by-product of the binary classification objective.

\subsection{Inference and XAI Module}
\label{sec:xai}

\noindent\textbf{Inference.}
Each CT volume $\mathcal{V}$ may possess a varying number $K'$ of axial slices, rendering each model trained on a specific volume constrained in its applicability. To address this issue, \textit{HexMIL} splits each volume into non-overlapping $K$-slice sub-volumes $\mathcal{V}_i, \quad i=1,\dots,T$, where $T=\texttt{int}(K'/K)$~\footnote{If $K' < K$ or $(K' \ne 0 \mod K)$, a padding strategy is used.} and assigns a probability score $p_i$ to each for tampering. The entire volume score is determined by the highest scores among all:

\begin{equation}
    \mathbb{P}(Y=1|\mathcal{V})=\max(p_1,\dots,p_T).
\end{equation}

\noindent\textbf{Hierarchical heatmap construction.}
At inference time, both attention maps are jointly available for each sub-volume: the inter-slice weights $\beta_k$ from the Volume Gated Attention and the intra-slice patch weights $\boldsymbol{\alpha}_k = [\alpha_{k,1}, \dots, \alpha_{k,N}]^\top \in \mathbb{R}^N$ from the Patch Gated Attention. Two thresholds govern their combination and operate at \emph{different stages} of the pipeline on \emph{different signals}: $\tau_\beta$ filters individual slices based on the raw inter-slice weight $\beta_k$ \emph{before} the volume is formed, while $\tau_{3D}$ selects voxels based on the \emph{combined, normalized} product $\beta \times \alpha$ after the full 3D attention volume is assembled.

Concretely, for each slice $k$ we first gate out uninformative slices:
\begin{equation}
    \tilde{\beta}_k = \beta_k \cdot \mathbf{1}[\beta_k \geq \tau_\beta].
    \label{eq:beta_mask}
\end{equation}
Slices for which $\tilde{\beta}_k = 0$ contribute nothing to the heatmap; their intra-slice attention map $\boldsymbol{\alpha}_k$ is never evaluated. For the remaining slices, $\boldsymbol{\alpha}_k \in \mathbb{R}^N$ is reshaped into a 2D spatial grid and bilinearly upsampled to the original slice resolution, yielding $\tilde{A}_k \in \mathbb{R}^{H \times W}$. The raw local attention volume for sub-volume $\mathcal{V}_i$ is then assembled as:
\begin{equation}
    \mathcal{H}^{(i)}_{3D}[k, i, j] = \tilde{\beta}_k \cdot \tilde{A}_k[i,\, j],
    \quad k = 1,\ldots,K,\;\; (i,j) \in [H]\!\times\![W].
    \label{eq:heatmap}
\end{equation}
Each voxel encodes simultaneously \emph{where} within a slice the model focuses (via $\tilde{A}_k$) and \emph{which} slices are most informative for the volume decision (via $\tilde{\beta}_k$).

\begin{figure}[t!]
    \centering
    \includegraphics[width=\linewidth]{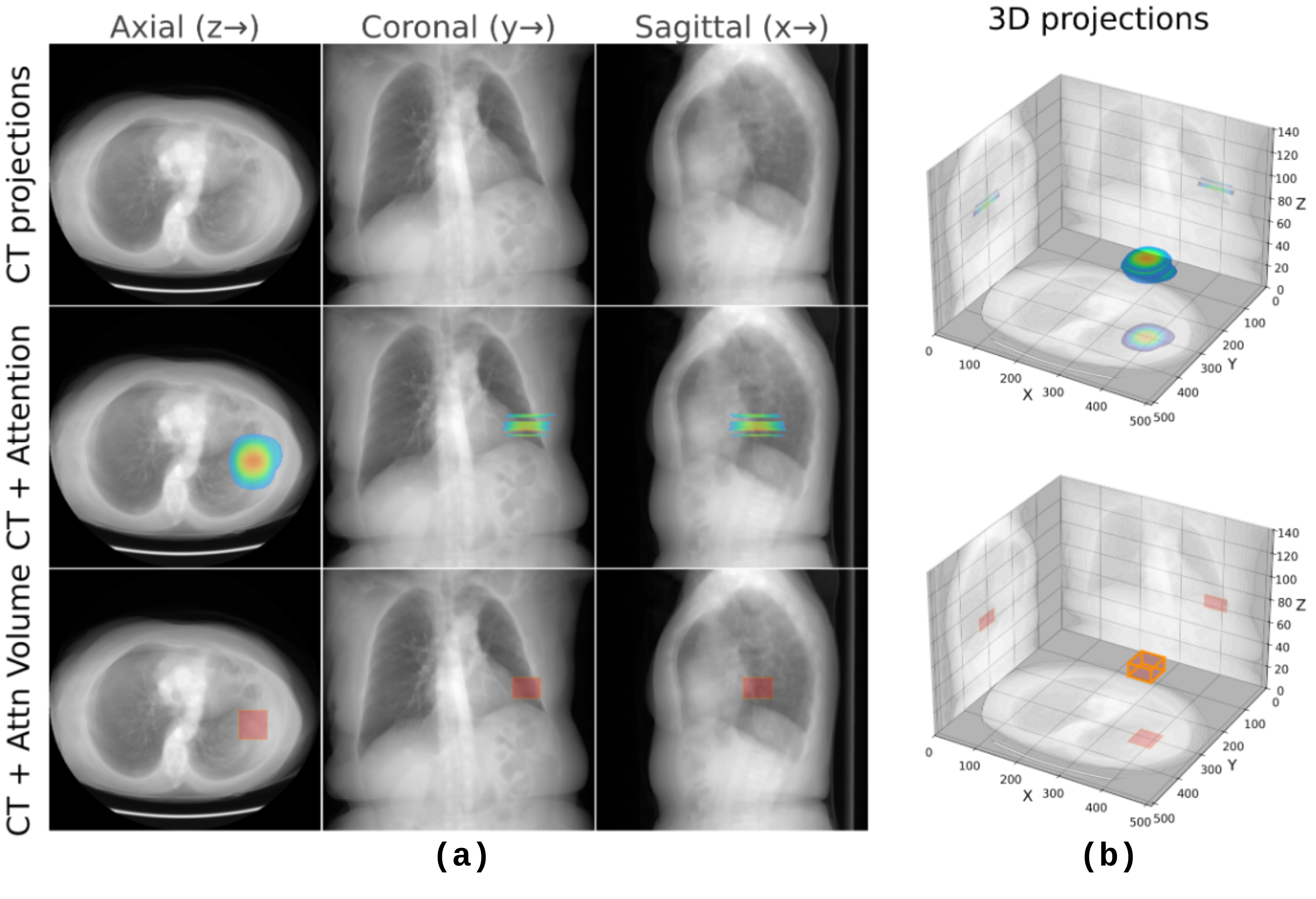}
    \caption{XAI Module output. (a) Projection of attention maps and extracted Attention Volume along three axes. (b) 3D visualization of the attention heatmap (top) and the resulting 3D bounding box for spatial localization (bottom).}
    \label{fig:bbox3d}
\end{figure}

\noindent\textbf{Full-volume heatmap assembly.}
The procedure above yields a raw local attention volume $\mathcal{H}^{(i)}_{3D} \in \mathbb{R}^{K \times H \times W}$ for each sub-volume $\mathcal{V}_i$. When $T > 1$, these are concatenated along the slice axis to reconstruct the full-resolution map:
\begin{equation}
    \mathcal{H}_{3D} = \left[\mathcal{H}^{(1)}_{3D} \;\big|\; \mathcal{H}^{(2)}_{3D} \;\big|\; \cdots \;\big|\; \mathcal{H}^{(T)}_{3D}\right] \in \mathbb{R}^{K' \times H \times W}.
    \label{eq:heatmap_concat}
\end{equation}
Smoothing and normalization are then applied \emph{globally} to $\mathcal{H}_{3D}$, so that relevance scores are comparable across the entire scan rather than within each sub-volume independently:
\begin{equation}
    \hat{\mathcal{H}}_{3D} =
        \frac{G_\sigma * \mathcal{H}_{3D} - \min(\cdot)}
             {\max(\cdot) - \min(\cdot) + \varepsilon},
    \label{eq:norm}
\end{equation}
where $G_\sigma$ is a Gaussian kernel. After normalization, a value of $\hat{\mathcal{H}}_{3D}[k,i,j] = v$ means that voxel $(k,i,j)$ ranks in the top $(1-v)$ fraction of the most attended voxels in the entire scan. \\

\noindent\textbf{Attention Volume extraction.}
We threshold $\hat{\mathcal{H}}_{3D}$ at $\tau_{3D}$ and extract the tightest axis-aligned 3D bounding box:
\begin{equation}
    \mathcal{M}_{\tau_{3D}} = \mathbf{1}\!\left[\hat{\mathcal{H}}_{3D} > \tau_{3D}\right],
    \qquad
    B_{\tau_{3D}} = \operatorname{BBox3D}(\mathcal{M}_{\tau_{3D}}),
    \label{eq:bbox}
\end{equation}
where $B_{\tau_{3D}} = (z_{\min}, z_{\max}, y_{\min}, y_{\max}, x_{\min}, x_{\max})$. Because $\hat{\mathcal{H}}_{3D}$ is globally normalized, $\tau_{3D}$ has a relative interpretation: $\tau_{3D} \in (0,1)$ selects the top $(1-\tau_{3D})\cdot100\%$ of the most attended voxels, regardless of the absolute magnitude of the attention scores. \Cref{fig:bbox3d} provides a visual example of the resulting projected maps and the final 3D bounding box.

\section{Experiments}
\label{sec:experiments}

\noindent\textbf{Datasets.}
We evaluate on two complementary benchmarks for local 3D manipulation detection in medical CT volumes, both built on the LIDC-IDRI collection~\cite{armato2011lung}.
\textit{CT-GAN}~\cite{mirsky2019ct} introduced the attack scenario of injecting and removing lung cancer nodules via a conditional GAN, producing a set of tampered volumes with pixel-level manipulation masks.
\textit{M3DSynth}~\cite{zingarini2024m3dsynth} extends this scenario to three diverse generative architectures --- a conditional pix2pix GAN, a CycleGAN, and a Diffusion Model --- each producing both injection and removal attacks, for a total of six manipulation configurations and 8{,}577 manipulated volumes alongside their pristine counterparts (over 11{,}000 nodule-level samples).
The three modalities exhibit markedly different artifact signatures, making cross-modality evaluation a rigorous proxy for out-of-distribution generalization.
To the best of our knowledge, M3DSynth and CT-GAN are the only publicly available benchmark for \emph{local} 3D manipulation detection in medical imaging. Both datasets adopt patient-level train/validation/test splits to prevent data leakage.

\noindent\textbf{Implementation details.}
The patch encoder $f_\theta$ is a ResNet-50~\cite{he2016deep}. Each axial slice is tiled into overlapping patches of size $P\!=\!64$ with stride $S\!=\!32$. Patch features are projected to $D\!=\!512$ dimensions and patch gated attention module uses an internal dimension $L\!=\!128$. Data augmentation consists of random horizontal and vertical flips. The frozen SliceMIL processes windows of $K\!=\!32$ consecutive slices, while slice gated attention uses an internal dimension $L'\!=\!256$. We train both SliceMIL and HexMIL for 100 epochs with AdamW ($\mathrm{lr}\!=\!10^{-4}$, $\mathrm{wd}\!=\!10^{-5}$). For the \textit{XAI Module} we use $\tau_\beta = 0.1$, $\sigma=0.03\max(K,H,W)$, and $\tau_{3D}=0.65$. All experiments use a single NVIDIA A6000 GPU and are fully reproducible.

\noindent\textbf{Baselines.}
We compare \textit{HexMIL} with an extensive set of baselines, including DeepFeatureX-SN~\cite{pontorno2025deepfeaturex}, FreqNet~\cite{tan2024frequency}, NPR~\cite{tan2024rethinking}, $D^3$~\cite{yang2025d}, Xception~\cite{chollet2017xception}, HP-FCN~\cite{li2019localization}, ManTraNet~\cite{wu2019mantra}, MVSS-Net~\cite{chen2021image}, and TruFor~\cite{guillaro2023trufor}.
Among these, DeepFeatureX-SN~\cite{pontorno2025deepfeaturex} uses contrastive learning to extract generator traces, while $D^3$~\cite{yang2025d} relies on feature discrepancy signals. To target structural and spectral artifacts, NPR~\cite{tan2024rethinking} models up-sampling patterns, whereas FreqNet~\cite{tan2024frequency} and HP-FCN~\cite{li2019localization} exploit high-frequency representations and high-pass filtering to expose inpainting traces.
All methods are inherently 2D: at inference, each axial slice is processed independently and the volume-level score is obtained by max-pooling across slice predictions.
Furthermore, we evaluate native 3D and spatiotemporal baselines that directly capture volumetric context: 3D CNNs (\textit{R3D-18}~\cite{tran2018closer}, \textit{DenseNet121-3D}~\cite{hara2018can}) and Video Vision Transformers (\textit{ViViT}~\cite{arnab2021vivit} in both Joint Space-Time and Factorised settings). Finally, \textit{ResNet50-ABMIL} and \textit{ViT-ABMIL} serve as single-level MIL controls, combining a ResNet-50~\cite{he2016deep} or ViT~\cite{dosovitskiy2020image} backbone with gated attention~\cite{ilse2018attention} applied slice-by-slice, without any volume-level aggregation stage. All baselines have been trained using their official repositories.

\subsection{Results}
\label{sec:res}
We evaluate \textit{HexMIL}'s capabilities in generalization for both classification and localization of the AI-manipulated volume. We train the model using the manipulated volumes from a single architecture present in the M3DSynth~\cite{zingarini2024m3dsynth} and CT-GAN~\cite{mirsky2019ct} datasets as fake parts in each scenario, while testing with all the others excluded.

\begin{table*}[t]
    \centering
    \caption{%
    Out-of-domain classification performance (AUC\% / F1\%). Comparison of our proposed \textit{HexMIL} against state-of-the-art methods. Models are trained on a single manipulation type and tested on the others to evaluate cross-domain generalization. Best results are highlighted in bold, second-best are underlined.
    }
    
    \begin{adjustbox}{width=\textwidth, scale=2}
    \renewcommand{\arraystretch}{1.5}
    \begin{tabular}{L CCCCCCCCCCCC C}
        \toprule
        \multicolumn{14}{c}{\Huge \textbf{Out-of-domain Classification (AUC\,\% / F1\,\%)}} \\ 
        \midrule
        Train set ($\to$) & \multicolumn{3}{c}{{\huge Pix2Pix}} & \multicolumn{3}{c}{{\huge CycleGAN}} & \multicolumn{3}{c}{{\huge DM}} & \multicolumn{3}{c}{{\huge CT-GAN}} & \multirow{2}{*}{Avg} \\
        Test set ($\to$) & CycleGAN & DM & CT-GAN & Pix2Pix & DM & CT-GAN & Pix2Pix & CycleGAN & CT-GAN & Pix2Pix & CycleGAN & DM & \\ \cmidrule(lr){2-4} \cmidrule(lr){5-7} \cmidrule(lr){8-10} \cmidrule(lr){11-13} \cmidrule{14-14}

        R3D-18 \cite{tran2018closer}
            & 66.8/67.0 & 66.5/64.4 & 62.1/65.4
            & 70.8/64.0 & 70.1/67.2 & 64.3/61.8
            & 69.5/64.6 & 67.6/68.4 & 71.2/74.9
            & 60.4/66.2 & 61.3/62.1 & 65.5/68.0
            & 66.3/66.2 \\

        ViViT (Joint ST) \cite{arnab2021vivit}
            & 73.5/64.6 & 70.7/67.0 & 64.7/66.0
            & 69.8/68.2 & 72.4/70.4 & 73.6/71.0
            & 74.3/69.9 & 71.5/75.7 & 69.8/70.0
            & 72.2/68.7 & 71.7/60.6 & 70.9/64.7
            & 71.3/68.1 \\

        ViViT (Factorised) \cite{arnab2021vivit}
            & 74.2/69.2 & 70.9/76.1 & 76.9/74.3
            & 71.0/69.3 & 72.9/70.3 & 70.3/70.8
            & 72.4/70.9 & 73.1/71.0 & 69.8/74.5
            & 76.2/72.4 & 74.7/69.5 & 72.7/73.4
            & 72.9/71.8 \\

        DenseNet121-3D \cite{hara2018can}
            & 68.7/71.2 & 70.2/68.4 & 69.3/72.0
            & 68.7/62.8 & 68.3/64.5 & 71.1/70.5
            & 67.4/63.1 & 65.4/66.3 & 72.3/69.9
            & 70.2/73.0 & 69.9/66.1 & 71.6/72.9
            & 69.4/68.4 \\
        
        ResNet50-ABMIL \cite{he2016deep,ilse2018attention}
            & 64.6/62.8 & 77.6/70.2 & 65.8/66.0
            & 69.4/62.2 & 61.1/58.3 & 68.0/72.3
            & 61.9/58.4 & 54.3/59.6 & 57.8/61.0
            & 60.7/59.8 & 70.1/65.6 & 68.0/67.9
            & 65.3/63.7 \\

        ViT-ABMIL \cite{dosovitskiy2020image,ilse2018attention}
            & 65.3/70.6 & 68.5/68.0 & 61.2/66.4
            & 68.3/64.4 & 68.2/66.7 & 64.9/62.1
            & 66.8/64.5 & 64.7/68.5 & 69.1/73.3
            & 58.7/65.0 & 62.4/60.5 & 60.1/68.8
            & 64.9/66.6 \\

        DFX-SN~\cite{pontorno2025deepfeaturex}
            & 73.1/\underline{\Huge 76.0} & 70.9/70.6 & 68.4/74.2
            & 63.4/63.1 & 75.0/65.8 & 59.7/61.3
            & 67.4/68.5 & 69.7/69.9 & 72.1/70.4
            & 66.2/71.5 & 64.8/63.4 & 67.9/75.8
            & 68.2/70.0 \\

        HP-FCN \cite{li2019localization}
            & 60.4/65.4 & 74.0/67.6 & 55.2/63.8
            & 64.5/60.3 & 61.1/63.1 & 67.8/65.4
            & 72.1/60.4 & 55.0/65.2 & 64.3/71.2
            & 53.4/59.1 & 66.2/74.4 & 57.9/56.6
            & 62.1/64.4 \\

        MVSS-Net$^\dagger$ \cite{chen2021image}
            & 63.4/66.9 & 72.9/63.2 & 58.7/64.2
            & 65.4/59.1 & 64.9/63.1 & 69.2/71.5
            & 74.8/58.9 & 59.0/66.8 & 67.5/65.9
            & 55.1/62.3 & 68.4/70.1 & 60.8/63.4
            & 65.0/64.6 \\

        FreqNet~\cite{tan2024frequency}
            & 62.0/71.4 & 71.9/67.4 & 57.8/63.2
            & 53.4/63.1 & 62.0/64.7 & 55.6/62.1
            & 72.4/75.5 & 68.7/65.0 & 66.5/72.3
            & 58.9/66.4 & 64.2/67.8 & 61.3/60.9
            & 62.9/66.6 \\

        NPR~\cite{tan2024rethinking}
            & 70.4/71.0 & 64.9/70.9 & 66.8/72.4
            & 77.4/73.1 & 79.0/65.3 & 71.2/70.5
            & 90.4/85.1 & 75.7/\underline{\Huge 76.9} & 84.1/88.3
            & 65.5/70.1 & 62.4/68.0 & 73.0/71.6
            & 73.4/73.6 \\

        TruFor \cite{guillaro2023trufor}
            & 84.0/75.9 & 84.9/70.4 & 71.2/\underline{\Huge 76.8}
            & 83.4/63.1 & 82.0/65.6 & 78.5/82.4
            & 90.4/65.5 & 78.7/74.9 & 85.1/\underline{\Huge 89.3}
            & 71.4/74.2 & 81.3/79.5 & 72.8/75.0
            & 80.3/73.3 \\

        D$^3$~\cite{yang2025d}
            & 75.0/\underline{\Huge 76.0} & 84.9/\underline{\Huge 75.4} & 68.4/74.2
            & 88.0/\underline{\Huge 87.1} & \underline{\Huge 92.2}/\underline{\Huge 82.6} & \underline{\Huge 83.2}/81.5
            & 89.4/\underline{\Huge 90.5} & \underline{\Huge 79.7}/\underline{\Huge 76.9} & 85.7/\textbf{\Huge 93.3}
            & 72.1/\underline{\Huge 78.9} & 79.4/77.2 & 70.3/\underline{\Huge 76.5}
            & 80.7/\underline{\Huge 80.8} \\

        ManTraNet \cite{wu2019mantra}
            & \underline{\Huge 86.1}/75.3 & \underline{\Huge 88.9}/63.3 & \underline{\Huge 73.5}/70.2
            & \underline{\Huge 89.0}/61.1 & 85.7/64.7 & 81.2/\underline{\Huge 88.4}
            & \underline{\Huge 93.4}/62.6 & 76.8/76.5 & \underline{\Huge 87.9}/85.3
            & \underline{\Huge 72.4}/77.1 & \underline{\Huge 83.5}/\underline{\Huge 81.2} & \underline{\Huge 75.8}/73.5
            & \underline{\Huge 82.9}/73.7 \\

        \rowcolor{mygray}
        \textbf{HexMIL (Ours)}
            & \textbf{\Huge 91.6}/\textbf{\Huge 86.2} & \textbf{\Huge 99.3}/\textbf{\Huge 97.5} & \textbf{\Huge 88.4}/\textbf{\Huge 92.1}
            & \textbf{\Huge 97.1}/\textbf{\Huge 93.5} & \textbf{\Huge 98.3}/\textbf{\Huge 96.0} & \textbf{\Huge 94.2}/\textbf{\Huge 91.8}
            & \textbf{\Huge 98.7}/\textbf{\Huge 96.5} & \textbf{\Huge 90.1}/\textbf{\Huge 80.8} & \textbf{\Huge 90.5}/84.2
            & \textbf{\Huge 85.1}/\textbf{\Huge 90.3} & \textbf{\Huge 86.4}/\textbf{\Huge 83.0} & \textbf{\Huge 84.2}/\textbf{\Huge 90.0}
            & \textbf{\Huge 92.0}/\textbf{\Huge 90.2} \\

        \bottomrule
    \end{tabular}
    \end{adjustbox}
    \label{tab:ood_cls}
\end{table*}

\begin{table*}[t]
    \centering
    \caption{%
    Out-of-domain localization performance (IoU\% / PG\%). Evaluation of spatial grounding robustness on unseen manipulation types. \textit{HexMIL} consistently outperforms baseline approaches in cross-domain scenarios.
    }
    
    \begin{adjustbox}{width=\textwidth}
    \renewcommand{\arraystretch}{1.5}
    \begin{tabular}{L CCCCCCCCCCCC C}
        \toprule
        \multicolumn{14}{c}{\textbf{\Huge Out-of-Domain Localization (IoU\,\% / PG\,\%)}} \\
        \midrule
        Train set ($\to$) & \multicolumn{3}{c}{{\huge Pix2Pix}} & \multicolumn{3}{c}{{\huge CycleGAN}} & \multicolumn{3}{c}{{\huge DM}} & \multicolumn{3}{c}{{\huge CT-GAN}} & \multirow{2}{*}{Avg} \\
        Test  ($\to$) & CycleGAN & DM & CT-GAN & Pix2Pix & DM & CT-GAN & Pix2Pix & CycleGAN & CT-GAN & Pix2Pix & CycleGAN & DM & \\
        \cmidrule(lr){2-4} \cmidrule(lr){5-7} \cmidrule(lr){8-10} \cmidrule(lr){11-13} \cmidrule{14-14}

        HP-FCN \cite{li2019localization}
            & 14.9/20.8 & 23.1/32.3 & 16.9/21.5 
            & 26.4/34.0 & 8.8/17.3 & 25.3/29.8 
            & 35.9/31.2 & 7.2/13.4 & 34.3/28.8 
            & 11.5/16.3 & 22.7/33.8 & 16.5/16.8 
            & 20.3/24.7 \\

        MVSS-Net \cite{chen2021image}
            & \textbf{\Huge 41.1}/60.3 & 33.9/50.6 & \textbf{\Huge 38.5}/\underline{\Huge 63.1} 
            & 40.8/56.8 & 30.3/43.8 & 38.1/58.9 
            & 46.2/66.5 & \underline{\Huge 29.3}/45.4 & 46.1/67.4 
            & \textbf{\Huge 42.7}/56.2 & 33.1/49.9 & \underline{\Huge 37.3}/59.1 
            & \underline{\Huge 38.1}/56.5 \\

        TruFor \cite{guillaro2023trufor}
            & 34.4/62.8 & 35.1/63.2 & 33.7/57.8 
            & 37.8/71.0 & \underline{\Huge 35.5}/\underline{\Huge 65.6} & 38.2/69.2 
            & \textbf{\Huge 53.9}/80.7 & 19.5/\underline{\Huge 47.8} & \textbf{\Huge 49.7}/\textbf{\Huge 83.7} 
            & 31.4/60.3 & 37.7/62.9 & 32.4/58.7 
            & 36.6/65.3 \\

        ManTraNet \cite{wu2019mantra}
            & 38.2/\textbf{\Huge 67.2} & \underline{\Huge 43.7}/\underline{\Huge 69.0} & 33.7/\underline{\Huge 63.1} 
            & \underline{\Huge 41.4}/\underline{\Huge 71.9} & 33.9/63.7 & \underline{\Huge 42.4}/\textbf{\Huge 73.7} 
            & 50.4/\underline{\Huge 83.5} & 10.6/33.0 & 47.9/\underline{\Huge 81.9} 
            & 34.3/\textbf{\Huge 65.5} & \textbf{\Huge 44.5}/\underline{\Huge 71.2} & 34.9/\textbf{\Huge 66.0} 
            & 38.0/\underline{\Huge 67.5} \\

        \rowcolor{mygray}
        \textbf{HexMIL (Ours)}
            & \underline{\Huge 40.0}/\underline{\Huge 66.1} & \textbf{\Huge 46.7}/\textbf{\Huge 75.6} & \underline{\Huge 37.8}/\textbf{\Huge 63.4}
            & \textbf{\Huge 47.1}/\textbf{\Huge 77.0} & \textbf{\Huge 44.1}/\textbf{\Huge 70.7} & \textbf{\Huge 43.5}/\underline{\Huge 73.2}
            & \underline{\Huge 50.7}/\textbf{\Huge 84.8} & \textbf{\Huge 31.7}/\textbf{\Huge 55.5} & \underline{\Huge 49.2}/81.1
            & \underline{\Huge 36.4}/\underline{\Huge 62.5} & \underline{\Huge 42.1}/\textbf{\Huge 78.4} & \textbf{\Huge 38.9}/\underline{\Huge 64.0}
            & \textbf{\Huge 42.4}/\textbf{\Huge 70.6}\\

        \bottomrule
    \end{tabular}
    \end{adjustbox}
    \label{tab:ood_loc}
\end{table*}

\noindent\textbf{Out-of-domain classification across generators.}
\Cref{tab:ood_cls} shows the results obtained by \textit{HexMIL} and all the baselines in the various out-of-domain classification scenarios. The results show a clear hierarchy but also a significant structure among the strongest baselines. \textit{HexMIL} achieves the best overall classification performance with an AUC of $92.0$ and an F1 score of $90.2$, dominates all transfer directions in terms of AUC and nearly all for F1, which indicates robust cross-generator generalization rather than gains limited to a few favorable pairs. Relative to the strongest baseline averages, which reach $82.9$ AUC and $80.8$ F1, this translates to absolute increases of $+9.1$ and $+9.4$ (approximately $+10.9\%$ and $+11.6\%$ relative improvement, respectively). 
At the same time, baselines show distinct strengths: ManTraNet leads the competitors in average AUC with $82.9$, but its significantly lower average F1 of $73.7$ suggests a strong ranking ability offset by weak calibration under distribution shifts. $D^3$ attains the second-highest average AUC of $80.7$, yet its effectiveness varies across transfer directions, revealing a residual reliance on specific generative artifacts. NPR proves competitive in transfers from diffusion models but remains inconsistent in the most challenging cross-domain scenarios. Also specialized forensic detectors, such as TruFor and FreqNet, which use noise-sensitive fingerprints or high-frequency representations and have an average AUC score of $80.3$ and $62.9$, also have trouble when appearance and forensic statistics change at the same time across unseen generators. The results also show that single-level attention is not enough. Models like ResNet50-ABMIL and ViT-ABMIL do not have a second hierarchical stage and only use gated attention at the slice level. As a result, they perform much worse, with an average AUC close to $65$.
Finally, 3D CNNs (\textit{R3D-18}, \textit{DenseNet121-3D}) and Video Transformers (\textit{ViViT}) underperform \textit{HexMIL} by a large margin, reaching at most $72.9\%$ AUC. This gap stems from the spatial sparsity of CT manipulations, as tampered sub-volumes occupy only a small fraction of the scan. While standard voxel-based architectures fail to isolate such sparse signals from healthy background tissue, \textit{HexMIL}'s hierarchical attention progressively focuses on the most discriminative patches and slices.\\
\begin{figure}[t!]
    \centering
    \includegraphics[width=\linewidth]{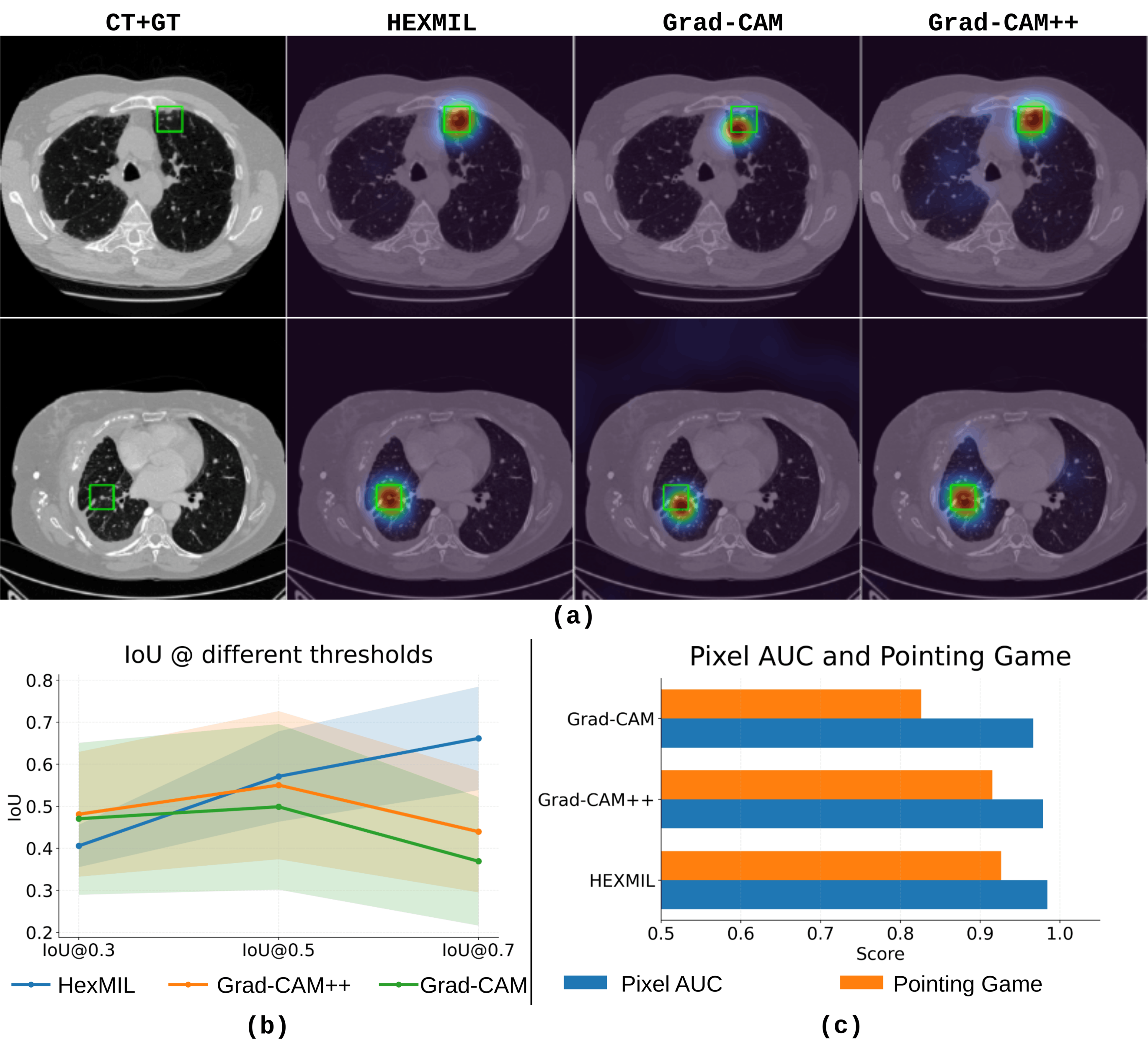}
    \caption{(a) Qualitative localization on manipulated CT slices (green boxes: ground truth), comparing our ante-hoc attention against post-hoc methods (Grad-CAM, Grad-CAM++) applied to the same \textit{HexMIL} model. (b) Quantitative evaluation using IoU. (c), Pixel AUC, and Pointing Game.
    }
    \label{fig:xai}
\end{figure}
\noindent\textbf{Out-of-domain localization results.}
The localization results in \Cref{tab:ood_loc} are particularly relevant because out-of-domain (ood) localization is generally more fragile compared to ood classification. Even in this context, \textit{HexMIL}, despite not being trained in the localization of the manipulated volume within CT scans, achieves the best average ood localization performance, with a Intersection-over-Union (IoU) $42.4$ and a Pointing Game (PG) $70.6$, and leads in most transfer cells, while the stronger baselines optimize different and often complementary behaviors. MVSS-Net and ManTraNet are almost equal in average IoU ($38.1$ vs. $38.0$), but ManTraNet records the highest baseline PG of $67.5$, consistent with sharper anomaly peaks; TruFor remains competitive with an average of $36.6$ IoU and $65.3$ PG, performing strongly in specific diffusion-to-GAN transfers yet degrading elsewhere due to its sensitivity to joint changes in appearance and forensic residual statistics. Taken together, these results suggest that previous methods are typically strong along one axis at a time (classification, calibration, overlap, or pointing), while \textit{HexMIL} is consistently strong across all axes. A plausible explanation is architectural: instead of a 2D slice-level evaluation followed by a late aggregation, \textit{HexMIL} performs a 3D hierarchical aggregation of evidence with ante-hoc attention, preserving the inter-slice context and producing localization maps that are directly linked to the prediction mechanism.

\subsection{Analysis}
\noindent\textbf{Interpretability.}
We evaluate \textit{HexMIL}'s intrinsic spatial explanations against two established post-hoc baselines (Grad-CAM~\cite{selvaraju2017grad}, Grad-CAM++~\cite{chattopadhay2018grad}) applied to the same ResNet-50 backbone. As shown in \Cref{fig:xai}~(a) and (b), all methods successfully identify manipulated regions in-domain, with \textit{HexMIL}'s forward attention weights offering comparable or slightly superior spatial grounding (98.4\% Pixel AUC, 92.6\% PG) relative to Grad-CAM (96.6\% AUC, 82.6\% PG) and Grad-CAM++ (97.8\% AUC, 91.5\% PG), while producing more compact heatmaps around the target sub-volume. Crucially, because \textit{HexMIL}'s attribution stems directly from its classification mechanics rather than post-hoc gradient backpropagation, its attention maps preserve higher spatial fidelity under out-of-domain distribution shifts, highlighting the structural reliability of ante-hoc explanations.

\noindent\textbf{Ablation Studies.}
We ablate \textit{HexMIL} along three axes: architectural components (Table~\ref{tab:abl_components}), attention mechanism (Table~\ref{tab:abl_attention}), and patch-encoder design (Table~\ref{tab:abl_stage1}). All results are reported on the out-of-domain test scenario. Two internal baselines serve as anchors throughout. \textit{ResNet50-ABMIL} applies a ResNet-50 slice-by-slice with global average pooling and max-pools per-slice scores into a single volume prediction — no MIL, no hierarchy. \textit{Pool-MIL} is structurally identical to \textit{HexMIL} but replaces gated attention with mean pooling at \emph{both} aggregation levels, isolating the attention contribution from the two-stage design.\\
\noindent\textit{Architectural components (Table~\ref{tab:abl_components}).}
The two-stage MIL hierarchy is the single largest contributor: switching from Flat-CNN to Pool-MIL alone yields $+9.6$ AUC and $+13.3$ F1, confirming that the ability to selectively weight patches within a slice and slices within a volume is critical, independently of whether the aggregation uses attention or pooling. Introducing gated attention adds a further $+11.8$ AUC, demonstrating that mean pooling under-exploits the non-uniform distribution of manipulation artifacts across slices. Sinusoidal positional encoding contributes an additional $+5.3$ AUC, which we attribute to the model learning that certain Z-ranges are anatomically more susceptible to specific manipulation patterns.

\begin{table}[t]
    \centering
     \caption{%
    Architectural ablation. Impact of progressively adding key components on out-of-domain classification.   \colorbox{mygray}{Grey}: selected configuration.
    }
    \label{tab:abl_components}
    
    \begin{adjustbox}{width=\columnwidth}
    \renewcommand{\arraystretch}{0.9}
    \newcommand{\cmark}{\textcolor{green!55!black}{$\checkmark$}}%
    \newcommand{\xmark}{\textcolor{red!70!black}{$\times$}}%
    \begin{tabular}{l cccc cc}
        \toprule
        Model & MIL & 2-stage & Gated & Pos.\,Enc. & Ood AUC\,(\%) & Ood F1\,(\%) \\
        \midrule
        Flat-CNN         & \xmark & \xmark & \xmark & \xmark & 65.3 & 63.7 \\
        Pool-MIL         & \cmark & \cmark & \xmark & \xmark & 74.9 & 77.0 \\
        HexMIL (No PE)   & \cmark & \cmark & \cmark & \xmark & 86.7 & 84.1 \\
        \rowcolor{mygray}
        \textbf{HexMIL}  & \cmark & \cmark & \cmark & \cmark & 92.0 & 90.2 \\
        \bottomrule
    \end{tabular}
    \end{adjustbox}
\end{table}
\vfill
\begin{table}[h]
    \centering
    \caption{
    Attention mechanism ablation. Comparison of different attention types on classification and localization performance.
    \colorbox{mygray}{Grey}: selected configuration.   }
    \label{tab:abl_attention}
    
    \renewcommand{\arraystretch}{0.7}
    \begin{tabular}{l cc cc}
        \toprule
        & \multicolumn{2}{c}{Ood Classification} & \multicolumn{2}{c}{Ood Localization} \\
        \cmidrule(lr){2-3} \cmidrule(lr){4-5}
        Attention type & AUC\,(\%) & F1\,(\%) & IoU\,(\%) & PG\,(\%) \\
        \midrule
        Standard        & 76.4 & 74.0 & 35.7 & 49.1 \\
        Self-Attention  & 93.5 & 91.2 & 1.6 &  6.0  \\
        \rowcolor{mygray}
        \textbf{Gated}   & \textbf{92.0} & \textbf{90.2} & \textbf{42.4} & \textbf{70.6} \\
        \bottomrule
    \end{tabular}
\end{table}

\begin{figure}[t!]
    \centering
    \includegraphics[width=\columnwidth]{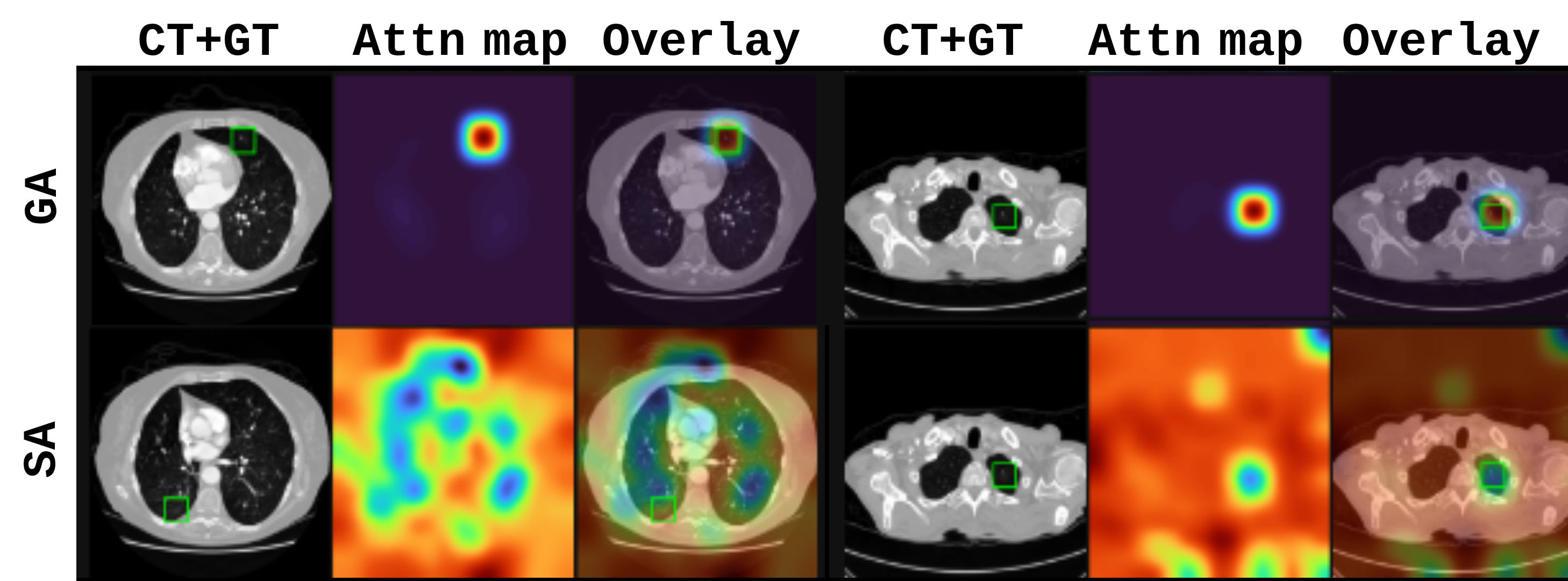}
    \caption{Visual comparison of Gated (GA, top) and Self-Attention (SA, bottom) maps. Green boxes denote ground truth manipulations.}
    \label{fig:ga_vs_sa}
\end{figure}

\noindent\textit{Attention mechanism (Table~\ref{tab:abl_attention}).}
We compare three aggregation strategies keeping backbone, patch size and window size fixed ($P\!=\!64$, ResNet-50, $K\!=\!32$). Standard attention~\cite{bahdanau2014neural} performs poorly on both classification ($76.4$ AUC) and localization ($35.7$ IoU, $49.1$ PG), suggesting that the sigmoid gating in \Cref{eq:alpha} is essential to suppress the majority of uninformative patches that dominate CT volumes.
Self-attention nearly matches gated attention on classification ($93.5$ vs.\ $92.0$ AUC) but suffers a catastrophic collapse on localization: IoU drops from $42.4\,\%$ to $1.6\,\%$ and the Pointing Game score from $70.6\,\%$ to $6.0\,\%$. This is a direct consequence of the global key--query mixing in self-attention: once patch representations are blended across the entire sequence, the one-to-one correspondence between attention weights and spatial locations is destroyed, and the resulting heatmap $\hat{\mathcal{H}}_{3D}$ loses all geometric meaning (\Cref{fig:ga_vs_sa}). Gated attention is the only mechanism that achieves competitive classification and interpretable localization simultaneously, precisely because each weight $\alpha_n$ (resp.\ $\beta_k$) retains an unambiguous spatial identity.  

\begin{table}[t!]
    \centering
    \caption{%
     Backbone and hyperparameter ablation. Out-of-domain performance across different feature extractors and values of $P$. Grey denotes the selected configuration. \colorbox{mygray}{Grey}: selected configuration.
    }
    \label{tab:abl_stage1}
    
    \begin{adjustbox}{width=\columnwidth}
    \renewcommand{\arraystretch}{0.7}
    \begin{tabular}{l cc cccc}
        \toprule
        & & \multicolumn{2}{c}{Ood Classification} & \multicolumn{2}{c}{Ood Localization} \\
        \cmidrule(lr){3-4} \cmidrule(lr){5-6}
        Backbone & $P$ & AUC\,(\%) & F1\,(\%) & IoU\,(\%) & PG\,(\%)\\
        \midrule
        DenseNet-121    & 32  & 84.2 & 87.0 & 43.2 & 73.4 \\
        DenseNet-121    & 64  & 90.9 & 88.3 & 40.8 & 70.1 \\
        DenseNet-121    & 128 & 91.1 & 89.0 & 28.4 & 67.6 \\
        \midrule
        EfficientNet-B0 & 32  & 81.6 & 85.2 & 42.3 & 62.1 \\
        EfficientNet-B0 & 64  & 87.2 & 88.9 & 39.0 & 59.4 \\
        EfficientNet-B0 & 128 & 89.1 & 86.4 & 36.9 & 50.0 \\
        \midrule
        ResNet-50       & 32  & 88.2 & 90.0 & 44.1 & 69.2 \\
        \rowcolor{mygray}
        \textbf{ResNet-50} & \textbf{64} & \textbf{92.0} & \textbf{90.2} & \textbf{42.4} & \textbf{70.6} \\
        ResNet-50      & 128 & 93.1 & 90.9 & 37.2 & 68.5 \\
        \bottomrule
    \end{tabular}
    \end{adjustbox}
\end{table}

\noindent\textit{Patch-encoder design (Table~\ref{tab:abl_stage1}).}
ResNet-50 is the best backbone at every patch size, outperforming DenseNet-121 and EfficientNet-B0 by up to $4.8$ AUC ($92.0$ vs.\ $87.2$ at $P\!=\!64$). The patch size reveals a fundamental trade-off between classification and localization: $P\!=\!128$ attains the highest OOD AUC ($93.1$) but the worst IoU ($37.2\,\%$), while $P\!=\!32$ improves IoU ($44.1\,\%$) at the cost of classification ($88.2$ AUC). We select $P\!=\!64$ as the operating point that best balances the two objectives ($92.0$ AUC, $42.4\,\%$ IoU). The degradation in localization at large $P$ is expected: coarser patches produce a sparser attention grid, reducing the spatial resolution of $\hat{\mathcal{H}}_{3D}$ and making it harder to cover compact manipulation regions precisely.\\
\noindent Additional analyses and extended results are provided in the \emph{supplementary material}.

\section{Conclusion \& Limitations}
\label{sec:conclusion}
We presented \textit{HexMIL}, a hierarchical MIL framework that jointly addresses out-of-domain generalization and ante-hoc 3D spatial attribution for CT manipulation detection, trained with binary volume-level labels only. Across a rigorous cross-generator generalization protocol, \textit{HexMIL} outperforms all baselines by $+9.1$ AUC and $+9.4$ F1 in classification and achieves the best average IoU and PG in localization without any spatial supervision.\\
\textbf{Limitations.} Fixed-length windowing may bisect manipulation regions in long scans; an overlapping or adaptive strategy could mitigate this. Generalization to anatomical sites beyond the thorax and to other modalities (e.g., MRI) remains to be verified; the underlying data originates from the LIDC-IDRI collection. Future work must assess the robustness of the framework against multi-center data and variations in CT acquisition protocols, which are known to introduce significant domain shifts in medical imaging tasks.

\section*{Acknowledgments}
Orazio Pontorno is a PhD candidate enrolled in the National PhD in Artificial Intelligence, XXXIX cycle, organized by Università Campus Bio-Medico di Roma.
This work was supported by the DEFORM project, funded by the European Union's Horizon Europe Research and Innovation Programme under Grant Agreement No. 101308502.

\bibliographystyle{unsrt}
\bibliography{main}

\end{document}